\documentclass[10pt,twocolumn,letterpaper]{article}

\usepackage[pagenumbers]{cvpr} 

\usepackage{graphicx}
\usepackage{amsmath}
\usepackage{amssymb}
\usepackage{booktabs}

\usepackage{color}

\usepackage[inline]{enumitem}
\usepackage{array,multirow}
\usepackage{siunitx}
\usepackage{pifont}
\usepackage{textcomp}

\usepackage{tablefootnote}
\usepackage{makecell}

\usepackage[pagebackref,breaklinks,colorlinks]{hyperref}

\usepackage[capitalize]{cleveref}
\crefname{section}{Sec.}{Secs.}
\Crefname{section}{Section}{Sections}
\Crefname{table}{Table}{Tables}
\crefname{table}{Tab.}{Tabs.}

\def\cvprPaperID{5183} 
\def\confName{CVPR}
\def\confYear{2023}

\begin{document}

\title{Divide and Conquer: Answering Questions with Object Factorization and Compositional Reasoning}

\author{Shi Chen \qquad Qi Zhao\\
Department of Computer Science and Engineering,\\
University of Minnesota\\
{\tt\small \{chen4595, qzhao\}@umn.edu}}

\maketitle

\begin{abstract}
Humans have the innate capability to answer diverse questions, which is rooted in the natural ability to correlate different concepts based on their semantic relationships and decompose difficult problems into sub-tasks. On the contrary, existing visual reasoning methods assume training samples that capture every possible object and reasoning problem, and rely on black-boxed models that commonly exploit statistical priors. They have yet to develop the capability to address novel objects or spurious biases in real-world scenarios, and also fall short of interpreting the rationales behind their decisions. Inspired by humans' reasoning of the visual world, we tackle the aforementioned challenges from a compositional perspective, and propose an integral framework consisting of a principled object factorization method and a novel neural module network. Our factorization method decomposes objects based on their key characteristics, and automatically derives prototypes that represent a wide range of objects. With these prototypes encoding important semantics, the proposed network then correlates objects by measuring their similarity on a common semantic space and makes decisions with a compositional reasoning process. It is capable of answering questions with diverse objects regardless of their availability during training, and overcoming the issues of biased question-answer distributions. In addition to the enhanced generalizability, our framework also provides an interpretable interface for understanding the decision-making process of models. Our code is available at \url{https://github.com/szzexpoi/POEM}. 
\end{abstract}

\section{Introduction} \label{sec:intro}
One of the fundamental goals in artificial intelligence is to develop systems that are able to reason with the complexity of real-world data to make decisions. Most existing visual question answering (VQA) methods \cite{updown,mcb,ban,mlb,visualbert,oscar,vilbert,lxmert,mfb} assume a complete overlap between objects involved in training and testing, and commonly rely on the spurious distributions of questions and answers \cite{explicit_bias}. As a result, they have limited generalizability toward real-life visual reasoning, and also lack the ability to justify the reasoning process that leads to the answers. 

\textit{``All mammals are animals. All elephants are mammals. Therefore, all elephants are animals \cite{aristotle_logic}.''} The wide application of syllogistic logic reflects key characteristics of the ways humans reason about the world. Unlike models \cite{updown, vilbert, lxmert} that utilize implicit features and heavily exploit statistical priors, humans correlate diverse objects from the compositional perspective based on their shared characteristics \cite{deductive_reasoning} and tackle problems with a structured reasoning process, which is both generalizable and interpretable. 
\begin{figure*}
\centering
\includegraphics[width=1\linewidth]{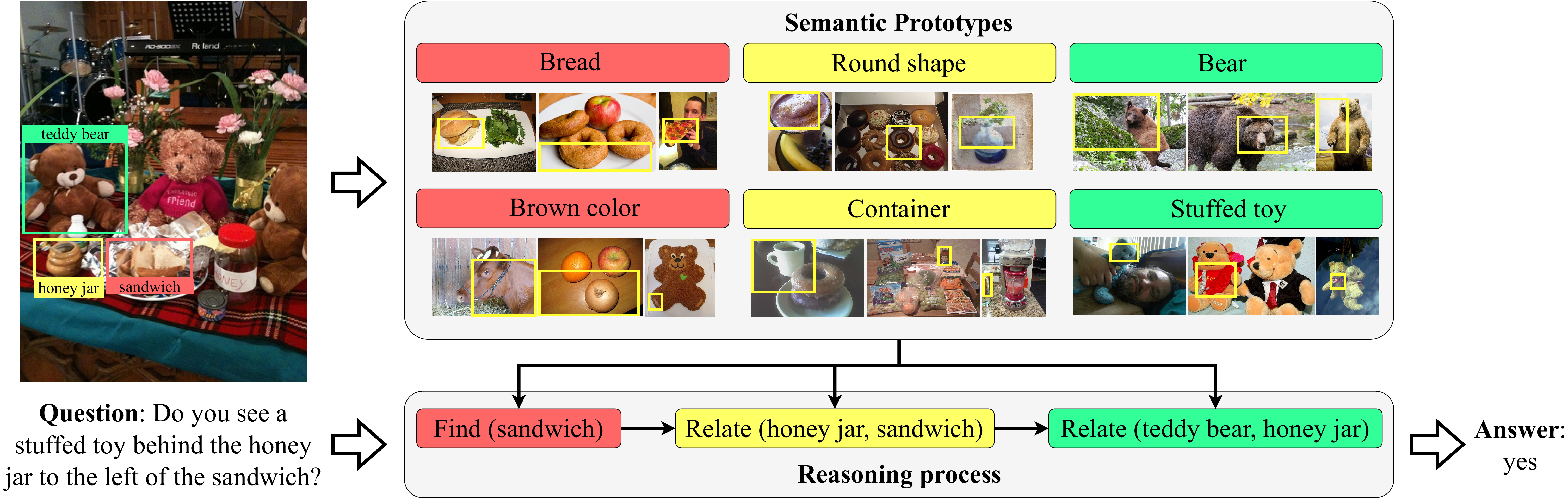}
\caption{Overview of our method that represents objects with semantically meaningful prototypes and makes decisions via an explicit reasoning process. Honey jar is a novel object unseen during training. Note that our prototypes are not limited to a set of manually defined categories, but learned from factorizing objects to encode broader characteristics (\textit{e.g.,} shapes, colors, object categories).}
\label{teaser}
\end{figure*}

To address the complexity of real-world problems, this study aims to develop object factorization and compositional reasoning capabilities in models. As shown in Figure \ref{teaser}, our approach bridges diverse objects by projecting them onto a common space formed by discriminative prototypes (\textit{e.g.}, round shape, stuffed toy), and formulates the reasoning process with atomic steps \cite{xnm} representing essential reasoning skills (\textit{e.g.}, \textit{Find}, \textit{Relate}). The prototypes are derived with object factorization, and they represent important semantics of objects (\textit{e.g.,} honey jar $\rightarrow$ $<$round shape, container ...$>$, teddy bear $\rightarrow$ $<$bear, stuffed toy ...$>$). With an improved understanding of semantic relationships, our framework correlates objects (\textit{e.g.,} honey jar and container, stuffed toy and teddy bear) based on their commonalities in characteristics, leading to enhanced robustness against the diversity of objects and data biases. It also allows interpretations of the model's reasoning process consistent with the ways humans describe their own thinking \cite{proto_fine_grained}. 

Compared to previous studies \cite{updown,stacknmn,visualbert,oscar,xnm,lxmert}, our major distinction lies in (1) the composition in two important dimensions of visual reasoning, \textit{i.e.}, objects and the reasoning process, and (2) a tight coupling between them. 
Instead of using black-boxed features or a direct mapping between question and answer that is vulnerable to object diversity or data biases, our method decomposes objects into bases representing discriminative semantics, and develops a prototypical neural module network to explicitly bridge objects with a compositional reasoning paradigm. 
The proposed method naturally approaches generalizability with its compositional nature, handling novel objects and variable data distributions. It also provides a transparent interface for interpreting how models parse objects based on their characteristics and incorporate them for visual reasoning.  


To summarize, our major contributions are as follows:
\begin{enumerate}
\item We identify the significance of tightly coupling the compositionality between objects and the reasoning process, and for the first time investigate its effectiveness in generalizable and interpretable reasoning.



\item We propose a principled method that automatically derives prototypes with object factorization, which plays a key role in encoding key characteristics of objects.

\item We develop a new neural module network that adaptively reasons on the commonalities of different objects along a structured decision-making process. 

\item We perform extensive analyses to shed light on the roles of compositionality in reasoning with novel objects and variable data distributions.
\end{enumerate}

\section{Related Works}
Our study is most related to previous efforts on visual question answering, zero-short learning for VQA, and VQA with out-of-distribution (OOD) questions. 

\textbf{Visual question answering.} With the diversity in language and visual semantics, visual question answering has become a popular task for studying models' reasoning capability \cite{vtt}. A large body of research develops datasets \cite{vqa1,scene_text_qa,vqa2,gqa,clevr,okvqa,visualcomet,vcr} and models \cite{updown,nmn,stacknmn,nsm,ban,visualbert,oscar,vilbert,xnm,lxmert} for VQA. Early datasets typically rely on crowd-sourcing \cite{vqa1,vqa2,vcr} to collect human-annotated questions. Several recent studies \cite{gqa,clevr,closure} use functional programs to automatically generate questions based on pre-defined rules and enable more balanced data distributions. There is also an increasing interest in investigating different types of reasoning, \textit{e.g.,} scene text understanding \cite{scene_text_qa}, reasoning on context \cite{visualcomet}, and knowledge-based reasoning \cite{okvqa}. These data efforts lead to the development of methods that advance VQA models from different perspectives, including multi-modal fusion \cite{mcb,mlb,mfb}, attention mechanism \cite{updown,ban}, structured inference \cite{stacknmn,nsm,mac,xnm}, and vision-and-language pretraining \cite{visualbert,oscar,vilbert,lxmert}. The aforementioned studies assume that every semantic in the test questions is well illustrated during training, and pay little attention to the models' generalizability to real-world scenarios that inevitably involve novel objects and diverse question-answer distributions. Our study aims to fill the gap with an integral framework that develops generalizable decision making capability with object factorization and compositional reasoning. 
\begin{figure*}
\centering
\includegraphics[width=1\linewidth]{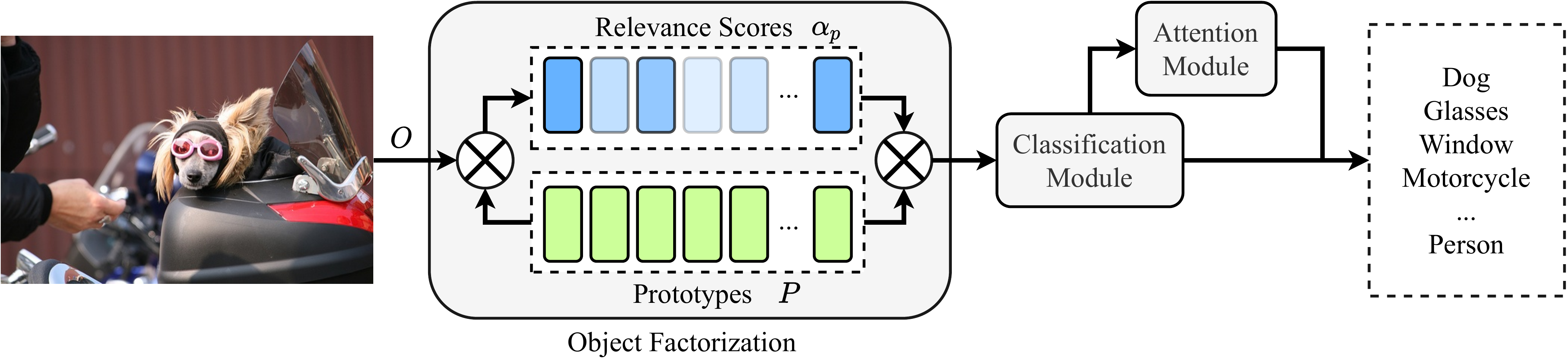}
\caption{Overview of our prototype learning paradigm based on object factorization.}
\label{proto_learning_fig}
\end{figure*}

\textbf{Zero-shot learning for VQA.} Zero-shot VQA aims to answer questions with novel objects. The pioneering work \cite{zero_shot_vqa_task} proposes the zero-shot setting for the VQA task, and benchmarks several techniques for improving the models' generalizability, including explicit object detector \cite{updown} and pretrained word embeddings \cite{glove}. Ramakrishnan \textit{et al.} \cite{zero_shot_empirical} leverage self-supervised pretraining to learn more generalizable features with external data. Wang \textit{et al.} \cite{vqa_machine} combine visual models trained on different datasets with an attention mechanism. Whitehead \textit{et al.} \cite{skill_concept} decompose VQA into two sub-problems, \textit{i.e.,} concept grounding and skill matching, and propose additional training objectives to address unseen objects in the questions. Besides the above studies concerning novel concepts in language, there are also attempts \cite{exemplar_att,vqa_meta_learning} that leverage exemplar-based methods to address novel concepts in both visual and textual modalities. While showing usefulness, these studies either leverage external data, which violates zero-shot assumption \cite{zero_shot_learning}, or require memorizing a vast amount of supporting samples. As a result, they have yet to develop the capability of generalizing toward real-world scenarios. The key differentiators of our method lie in its ability to connect novel and known objects and the use of an explicit reasoning process. By bridging objects based on their commonalities and decomposing the decision-making process into atomic reasoning steps, it improves the generalizability and interpretability without relying on external data or memorizing samples.


\textbf{VQA with OOD questions.} To develop models that can truly reason on the visual-textual inputs instead of relying on statistical priors, VQA with out-of-distribution questions has gained considerable attention. In particular, Agrawal \textit{et al.} \cite{vqa_cp} present the first VQA dataset with adversarial distributions between the training and validation data. A more recent study \cite{gqa_ood} analyzes models' robustness against biases by differentiating evaluation questions based on their question-answer distributions. To tackle the issues of harmful biases, a series of studies make progresses by improving visual attention \cite{hint,self_cirtical_vqa}, reducing biases toward individual modalities \cite{adv_reg, rubi, lmh_entropy}, and leveraging ensemble techniques \cite{lmh, grad_ensemble}. 
There are also attempts \cite{css, mutant, contrast_vqa} that use data augmentation to increase accuracy on biased data. However, they alter the distributions of training samples with additional data, and violate the original intent of the VQA with OOD questions task \cite{ood_shortcoming}. Our study is complementary to existing efforts, and it differentiates itself by investigating the usefulness of object factorization and compositional reasoning for addressing biases. It does not balance training samples to remove data biases, but instead focuses on enhancing models' own reasoning capabilities.

\section{Methodology}
Visual reasoning would benefit from capabilities of correlating objects based on their characteristics and decomposing problems into atomic steps \cite{ann_binding}. This section presents a new framework for improving the robustness against questions with novel objects and diverse question-answer distributions. It consists of two novel components: (1) a principled method that automatically derives semantically plausible prototypes to represent different objects, and (2) a neural module network that bridges objects by incorporating discriminative prototypes in an explicit reasoning process. Besides the enhanced generalizability, the method also provides an interpretable interface for elaborating on the rationales behind the model's decisions. 

\subsection{Bridging Diverse Objects with Factorization} \label{proto_learning_sec}
Inspired by humans' reasoning process that classifies objects based on their semantics similarity, a primary goal of our study is to derive semantic prototypes that can represent a vast amount of objects. The prototypes encode different aspects of the objects, and augment models with the capability to bridge diverse objects for more generalizable reasoning. Unlike previous studies \cite{att_proto,transferrable_proto} that construct prototypes based on manually annotations and have difficulties scaling to different scenarios, we propose to automatically learn discriminative prototypes by factorizing various objects. Object factorization plays a key role in parsing the fine-grained characteristics of objects (\textit{e.g.,} shapes, textures, and super-categories), which facilitates understanding of the semantic relationships between objects.
\begin{figure*}
\centering
\includegraphics[width=1\linewidth]{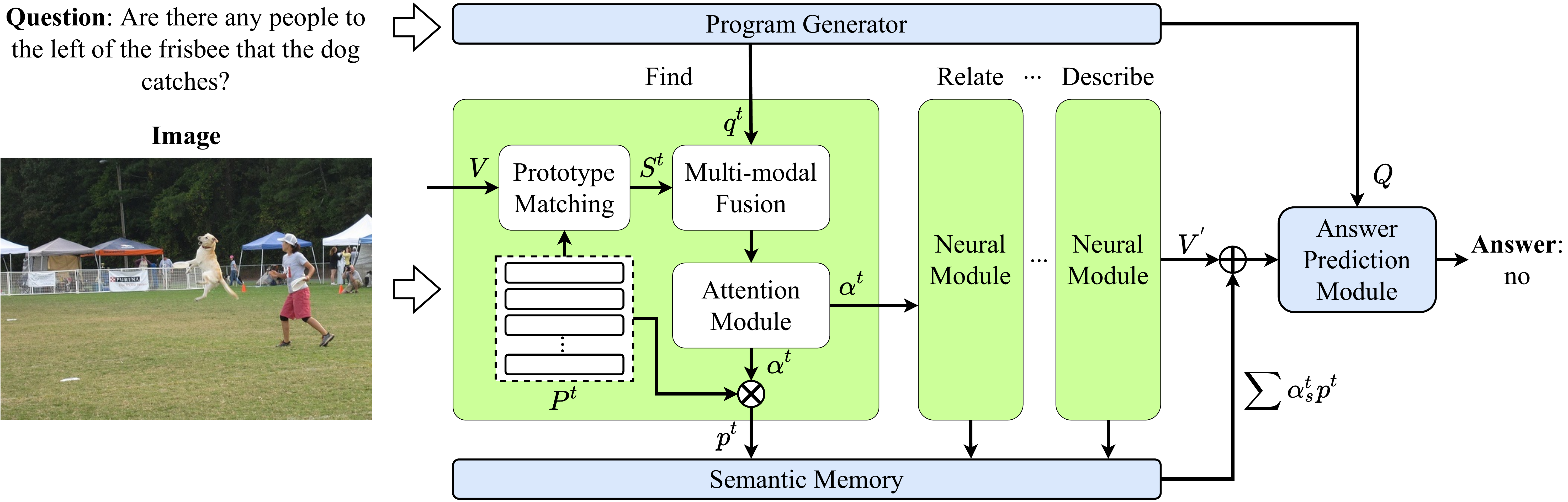}
\caption{Illustration of the proposed prototypical neural module network. $\otimes$ and $\oplus$ denote dot product and concatenation, respectively.}
\label{architecture}
\end{figure*}

As illustrated in Figure \ref{proto_learning_fig}, our prototype learning method leverages the multi-label classification task \cite{multi_label_survey,graph_multi_label,loss_multi_label,srn_multi_label} to discover discriminative prototypes in a data-driven manner. Given an input image, we train a deep neural network that predicts all object categories in the visual scene. Different from conventional approaches that recognize objects based on their visual features $O$, we decompose objects with trainable prototypes $P$ and utilize the combinations of their matched prototypes for classification:
\begin{equation} \label{sim_score}
    \alpha_{p} = \delta(O_{i} \cdot P)
\end{equation}
\begin{equation} \label{sim_score_k}
    C^{i} = Cls(\sum\limits_{k=1}^{K} \alpha_{p}^{k} P^{k})
\end{equation}
\noindent where $\alpha_{p} \in \mathbb{R}^{1 \times K}$ denotes relevance scores between the $i^{th}$ object and $K$ prototypes. $O_{i} \cdot P$ is the dot product between features and prototypes, which corresponds to their cosine similarity. $\delta$ is the sigmoid activation function for normalization. $C^{i}$ is the prediction for the current object and $Cls$ is the classification module. To facilitate optimization without expensive instance-level annotations, we leverage an attention module $A_{cls}$ to dynamically aggregate instance-level predictions into the image-wise prediction $C$:
\begin{equation}
    \alpha_{c}^{i} = A_{cls}(\sum\limits_{k=1}^{K} \alpha_{p}^{k} P^{k})
\end{equation}
\begin{equation}
    C = \sum\limits_{i=1}^{N} \alpha_{c}^{i} C^{i}
\end{equation}
\noindent where $N$ is the number of objects. We train the network with a standard binary cross-entropy loss, and select prototypes $P$ that provide the highest validation performance.

The aforementioned paradigm factorizes objects into a set of bases formed by different prototypes. The prototypes encode important semantics of objects (see Section \ref{proto_analysis}), and serve as critical components of our reasoning model, as detailed in the next subsection. 

\subsection{Prototypical Neural Module Network} \label{proto_net}

With our prototypes constructing the common semantic space for bridging diverse objects, we further propose a novel prototypical neural module network to adaptively incorporate the correlation between objects with a compositional reasoning process. Compared to previous VQA methods \cite{updown,exemplar_att,visualbert,zero_shot_empirical,xnm,lxmert,vqa_meta_learning,skill_concept}, the advantages of our model lie in (1) its capability to generalize to both known and novel objects in the visual-textual domains, (2) the enhanced robustness against the spurious data biases, and (3) the interpretability of the decision-making procedure. 

Figure \ref{architecture} provides an overview of the proposed neural module network with prototype matching and a semantic memory module. The principal idea behind our model is to take advantage of semantic relationships encoded in our learned prototypes (Section \ref{proto_learning_sec}), and reason with a structured decision-making procedure. Neural module networks \cite{nmn,stacknmn,xnm} are a body of interpretable reasoning methods that perform visual reasoning with two steps: (1) decomposing the reasoning process into discrete reasoning steps, where each step is associated with an atomic module (\textit{e.g.,} \textit{Find} module for locating regions of interest), and (2) sequentially executing the modules on visual-textual inputs and gathering information to predict the answer. In addition to the use of an explicit reasoning process, our model utilizes prototype matching to project features of diverse objects onto the semantic space formed by prototypes, enabling it to measure their semantic relationships and tightly couple relevant objects. A semantic memory module is also proposed to adaptively combine important semantics captured at different reasoning steps, which facilitates joint reasoning throughout the whole reasoning process. 

Specifically, unlike conventional methods that rely on raw visual features and pay little attention to semantic relationships between objects, our model leverages the learned prototypes to parse different objects and correlate them based on their fine-grained characteristics. At each $t^{th}$ reasoning step, our network computes the similarity between visual features and prototypes, explicitly representing objects based on their corresponding characteristics, and uses the similarity scores $S^{t}\in \mathbb{R}^{N\times K}$ for decision making:
\begin{equation}
    S_{i}^{t} = \phi(V_{i} \cdot P^{t})
\end{equation}
\noindent where $V_{i} \in \mathbb{R}^{1\times D}$ is $D$ dimensional visual features for the $i^{th}$ visual regions ($N$ regions in total), and $P^t \in \mathbb{R}^{D \times K}$ represents the $K$ prototypes. $V_{i} \cdot P^{t}$ is the dot product between features and prototypes. $\phi$ is the hyperbolic tangent activation function for normalizing the scores. Upon obtaining the similarity scores, we then locate the regions of interest for the current reasoning step: 
\begin{equation} \label{att_computation}
    \alpha^{t} = A^t(F^t(S^t,q^t))
\end{equation}
\noindent where $\alpha^{t} \in \mathbb{R}^{1\times N}$ is the attention map highlighting the important regions, $F^t$ and $A^t$ are the multi-modal fusion and attention module, respectively. $q^t$ is the query information derived from the question. 

Another key differentiator between our proposed model and existing neural module networks \cite{stacknmn,xnm} is the incorporation of semantic memory. Instead of determining the answer solely based on the question $Q$ and attended visual features $V^{'}$, we further take into account the prototypes attended over time $p^{t} = \alpha^{t} \cdot S^{t}$, and bridge objects of interest at different reasoning steps. Our semantic memory module uses an attention mechanism to adaptively incorporate key prototypes at different steps:
\begin{equation}
    \alpha_{s} = A_{s}(Q) 
\end{equation}
\begin{equation} \label{joint_combination}
    \hat{y} = Ans([V^{'};\sum\limits_{t=1}^{T}\alpha_{s}^{t} p^{t}],Q)
\end{equation}
\noindent where $\alpha_{s} \in \mathbb{R}^{1\times T}$ represents attention weights for $T$ reasoning steps, and $A_{s}$ is the module for attention computation. $\hat{y}$ is the predicted answer, and $Ans$ is the answer prediction module. $[;]$ denotes the concatenation of features.

Our proposed model associates objects based on their relationships with distinct prototypes, and adaptively combines important prototypes captured throughout the reasoning process. It takes advantage of the compositionality in both objects involved during visual reasoning (\textit{i.e.,} from objects to their characteristics) and the reasoning process (\textit{i.e.,} from questions to reasoning steps), which plays an essential role in addressing the diversity of objects and the spurious data biases. The compositional nature of the model also allows better interpretation of the underlying decision-making procedure (Section \ref{reasoning_analysis}). 

\section{Experiments}
This section presents implementation details (Section \ref{implementation}) and experiments to analyze the proposed method. We experiment with two different settings of VQA, including zero-shot VQA (Section \ref{zeroshot_result_sec}) and VQA with out-of-distribution questions (Section \ref{ood_result_sec}), to validate the robustness of our method. We also provide an ablation study with different prototypes (Section \ref{proto_ablation}) to demonstrate the advantages of object factorization. Besides examining the effectiveness of our method, we perform extensive analyses to shed light on the following questions: (1) What do prototypes learn? Do they encode common characteristics among objects? (Section \ref{proto_analysis}); and (2) How do models reason to answer diverse questions? (Section \ref{reasoning_analysis}) 

\subsection{Implementations} \label{implementation}
\textbf{Datasets.} The primary goal of our experiments is to study models' generalizability to tackle real-world problems. We experiment with two different settings, each representing a common type of generalization: (1) \textbf{Zero-shot VQA} estimates models' generalizability toward both known and novel objects. Three popular datasets are used in our experiments, including VQA \cite{vqa2}, GQA \cite{gqa} and the recently introduced Novel-VQA \cite{skill_concept}. Following \cite{zero_shot_empirical,skill_concept}, for the VQA and GQA datasets, we reconstruct their training and validation sets to have a disjoint set of objects. For each dataset, we randomly select ten objects from the object pools and use them as novel objects unavailable during training. Similar to \cite{exemplar_att,zero_shot_empirical,zero_shot_vqa_task}, we exclude all training questions that either contain words related to the selected objects or use images with the objects, and divide the original validation sets into Known and Novel splits based on objects required for reasoning. For the Novel-VQA dataset, we adopt the original training and test sets \cite{skill_concept}, which only considers novel objects in the questions (Novel-Q); (2) \textbf{VQA with OOD questions} focuses on evaluating models' generalizability to data with diverse question-answer distributions. We experiment with two popular datasets, including VQA-CP \cite{vqa_cp} with adversarial distributions between training and evaluation, and GQA-OOD \cite{gqa_ood} that utilizes balanced training questions but differentiates evaluation questions based on their question-answer distributions.

\textbf{Evaluation.} We evaluate models with common metrics for visual question answering. For the VQA, Novel-VQA, and VQA-CP datasets, we adopt VQA accuracy \cite{vqa1} as the evaluation metric, which considers multiple candidate answers. For the GQA and GQA-OOD datasets, we use the standard accuracy since each question has an unique answer. We also follow \cite{gqa_ood} and consider differences in accuracy when answering in-distribution and out-of-distribution questions (\textit{i.e.,} $\bigtriangleup$) on the GQA-OOD dataset.

\textbf{Model specification.} We use the state-of-the-art neural module network XNM \cite{xnm} as our baseline, which provides competitive performance on multiple datasets without loss of interpretability. We follow \cite{explicit_xnm} and replace the \textit{Transform} module with the \textit{Relate} module to improve attention propagation. Following \cite{oscar,lxmert,skill_concept}, we adopt the UpDown features \cite{updown} that capture 36 semantically meaningful regions (\textit{i.e.,} N=36) as the visual inputs. To enable understanding of unseen vocabulary, we follow \cite{zero_shot_vqa_task,skill_concept} and use Glove vectors \cite{glove} to initialize the word embeddings. Other settings, \textit{e.g.,} network specification and training configuration, are the same as those defined in the original papers \cite{xnm,explicit_xnm} without tuning. The aforementioned baseline is incorporated with our method discussed in Section \ref{proto_net} to enable object factorization and compositional reasoning.

\textbf{Prototype learning.} We derive our prototypes with the multi-label classification task \cite{multi_label_survey,srn_multi_label}, which aims to predict all object categories that exist in an image. The image-wise ground truth is constructed with object detection labels (VQA, VQA-CP and Novel-VQA) or scene graphs (GQA and GQA-OOD). UpDown features are used as inputs to the classification network, which encode semantic information of different objects. For zero-shot VQA, we train the model on each dataset with training images that do not contain the selected novel objects, and evaluate it on validation images with only known objects. For VQA with OOD questions, we use the original training and validation sets without excluding samples. We set the number of prototypes in our experiments to 1000 (\textit{i.e.,} $K=1000$), which is comparable to the number of fine-grained objects in GQA (\textit{i.e.,} $\sim 1000$ object categories). The network is trained with Adam \cite{adam} optimizer for 60 epochs, the learning rate and batch size are set to $4\times 10^{-4}$ and 128, respectively. Prototypes with the best validation performance are used in our VQA model. 
\begin{table}
\caption{Comparative results on zero-shot VQA.}
\centering
\resizebox{1\linewidth}{!}{
{\small
\begin{tabular}{cccccc}
\toprule
\multirow{2}{*}{} & \multicolumn{2}{c}{VQA} & \multicolumn{2}{c}{GQA} & Novel-VQA\\
\cmidrule(lr){2-3} \cmidrule(lr){4-5} \cmidrule(lr){6-6}
 & Known & Novel & Known & Novel & Novel-Q\\ 
\midrule
UpDown \cite{updown} & 55.65 & 48.53 & 52.73 & 51.35 & 51.40  \\
\midrule
VisualBert \cite{visualbert} & - & - & 59.85 & 58.80 & - \\
\midrule
Skill-Concept \cite{skill_concept} & - & - & - & - & 59.80 \\
\midrule
XNM \cite{xnm} & 62.05 & 52.81 & 59.39 & 57.54 & 57.54 \\
\midrule
XNM+POEM  & 63.80 & 54.82 & 60.60 & 59.71 & 60.73 \\
\bottomrule
\end{tabular}
}
}
\label{main_result}
\end{table} 

\subsection{Results on Zero-shot VQA} \label{zeroshot_result_sec}
We first demonstrate the effectiveness of our prototypical neural module (POEM) network on answering questions with both known and novel objects. We compare it with four approaches, including (1) UpDown \cite{updown} that does not explicitly model the reasoning process, (2) Our baseline XNM \cite{xnm} with an explicitly reasoning process, (3) VisualBert \cite{visualbert} that utilizes vision-and-language pretraining on external data containing the novel objects (\textit{i.e.,} MSCOCO \cite{mscoco}), and (4) Skill-Concept \cite{skill_concept} that is the current state-of-the-art on Novel-VQA dataset, which explicitly exploits novel objects in images with around 97$\%$ of them covered in a handcrafted reference training set. 
All of the models are trained with UpDown visual features and initialized with pretrained word embeddings.
\begin{table}
\caption{Comparative results on VQA with OOD questions.}
\centering
\resizebox{0.85\linewidth}{!}{
{\small
\begin{tabular}{cccc}
\toprule
\multirow{2}{*}{} & VQA-CP & \multicolumn{2}{c}{GQA-OOD}\\
\cmidrule(lr){2-2}\cmidrule(lr){3-4} 
 & acc $\uparrow$ & acc-tail $\uparrow$ & $\bigtriangleup$ $\downarrow$ \\ 
\midrule
Bias Product \cite{lmh} & 39.93 & 30.8 & 12.0 \\
\midrule
AdvReg \cite{adv_reg} & 41.17 &  -& - \\
\midrule
Hint \cite{hint} & 46.73 & -& - \\
\midrule
RUBi \cite{rubi} & 47.11 & 35.7 & 14.3 \\
\midrule
SCR \cite{self_cirtical_vqa} & 49.45 & - & - \\
\midrule
LMH \cite{lmh} & 52.05 & 32.2 & 11.5 \\
\midrule
LMH+Entropies \cite{lmh_entropy} & 54.55 & - & - \\
\midrule
XNM \cite{xnm} & 51.54 & 46.14 & 14.4 \\
\midrule
XNM+POEM & 53.99 & 46.89 & 10.7 \\
\bottomrule
\end{tabular}
}
}
\label{ood_result}
\end{table} 

We draw three key observations from the results in Table \ref{main_result}: (1) While Updown and XNM show similar accuracy under the standard VQA setting \cite{updown,xnm}, the latter provides stronger performance on zero-shot VQA. The results indicate that the compositional reasoning process is not only helpful in interpretability, but also important for model performance and generalizability; (2) By leveraging external data, VisualBert provides better performance than both aforementioned models. However, it violates the original intent of zero-shot VQA with novel objects actually covered in the external data, making them impractical for real-world scenarios with unseen objects; and (3) Differently, with our prototypes bridging different objects and the reasoning process, the proposed method improves the performance of the XNM baseline without relying on external data, and achieves overall the best results in answering questions with both known and novel objects in the visual-textual data. It also outperforms Skill-Concept on the Novel-VQA dataset that only considers novel objects in the questions. Note that Skill-Concept assumes the availability of novel objects in training images and is not applicable to zero-shot setting on VQA and GQA datasets, while our method has the capability to generalize toward broader scenarios.



\subsection{Results on VQA with OOD Questions} \label{ood_result_sec}
Next, we investigate the robustness of our method against spurious data biases. We compare our method with eight approaches that do not exploit additional data, including Bias Product \cite{lmh}, AdvReg \cite{adv_reg}, RUBi \cite{rubi}, and LMH+Entropies \cite{lmh_entropy} for reducing single-modal biases, Hint \cite{hint} and SCR \cite{self_cirtical_vqa} for boosting visual attention, LMH \cite{lmh} for learning the residual of biases, and our baseline XNM. Following \cite{lmh, lmh_entropy,grad_ensemble}, when experimenting on the VQA-CP dataset \cite{vqa_cp}, both XNM and our model incorporate the learned mixed-in module \cite{lmh} to address known biases.

Results in Table \ref{ood_result} show that our method is able to improve the XNM baseline by a considerable margin. With object factorization and compositional reasoning, it not only improves the robustness against adversarial distributions (\textit{i.e.,} VQA-CP), but also reduces the performance gap between answering in-distribution and out-of-distribution questions (\textit{i.e.,} $\bigtriangleup$ for GQA-OOD). Compared to existing methods that introduce additional regularization and have difficulties generalizing to different datasets \cite{rubi,lmh,lmh_entropy}, our method augments models' reasoning capability without imposing data-specific constrains and thus enjoys better generalizability. The aforementioned observations suggest the significance of explicitly bridging objects with their fine-grained characteristics for overcoming data biases. 

\subsection{Ablation Study on Object Factorization} \label{proto_ablation}
A key component of our framework is the proposed method for learning discriminative prototypes with object factorization. In this section, we perform an ablation study to investigate the usefulness of different prototypes (see our supplementary materials for additional ablation studies on model design). Specifically, we consider three types of alternative prototypes: (1) Prototypes that are randomly initialized and learned from scratch on the VQA task (Scratch); (2) Prototypes specific to manually defined objects (Object), which are learned with the same multi-label classification task as our approach but without object factorization; and (3) Prototypes derived from Glove \cite{glove} word embeddings of concepts covered in the Visual Genome \cite{visual_genome} dataset (Textual), including objects, attributes, and relationships.
\begin{table}
\caption{Comparative results for different prototypes.}
\centering
\resizebox{1\linewidth}{!}{
{\small
\begin{tabular}{ccccccc}
\toprule
\multirow{2}{*}{} & \multicolumn{2}{c}{VQA} & \multicolumn{2}{c}{GQA} & Novel-VQA & VQA-CP\\
\cmidrule(lr){2-3} \cmidrule(lr){4-5} \cmidrule(lr){6-6} \cmidrule(lr){7-7}
 & Known & Novel & Known & Novel & Novel-Q & OOD\\ 
\midrule
Scratch & 62.66 & 53.66 & 58.87 & 57.69 & 59.48 & 50.39\\
\midrule
Object & 62.56 & 53.23 & 60.48 & 59.46 & 56.67 & 52.23\\
\midrule
Textual & 60.61 & 52.33 & 55.31 & 53.22 & 59.17 & 51.09\\
\midrule
Ours & 63.80 & 54.82 & 60.60 & 59.71 & 60.73 & 53.99\\
\bottomrule
\end{tabular}
}
}
\label{ablation_proto}
\end{table} 

We made three observations from results in Table \ref{ablation_proto}: (1) Randomly initialized prototypes lead to inferior performance among all datasets, indicating the significance of explicitly learning semantically plausible prototypes; (2) While object-based prototypes show reasonable performance on the GQA dataset with detailed annotations (\textit{i.e.,} $\sim1000$ object categories), they have negligible improvements on the VQA, Novel-VQA, and VQA-CP datasets with abstract-level annotations (\textit{i.e.,} 80 object categories). The large gap in performance gain across datasets demonstrates the advantages of fine-grained categorization, and more importantly, highlights the need to learn discriminative prototypes without relying on extensive annotations. For this, our method utilizes object factorization to automatically decompose objects into more elaborated semantics, and brings considerable improvements among datasets with both abstract and detailed object annotations; (3) Textual prototypes result in a visible drop in accuracy, despite the consideration of various concepts. This is likely caused by the difficulty of correlating objects across the visual and textual domains. Differently, our method directly captures diverse characteristics of objects from visual data and does not suffer from the discrepancies between modalities.

\subsection{What Do Prototypes Learn? Do They Encode Common Characteristics among Objects?} \label{proto_analysis}
Results in the previous sections demonstrate that our method learns discriminative prototypes to represent a variety of objects and bring enhanced generalizability across various settings. This section further demonstrates its effectiveness by investigating how our prototypes correlate different objects. 
\begin{figure}
\centering
\includegraphics[width=1\linewidth]{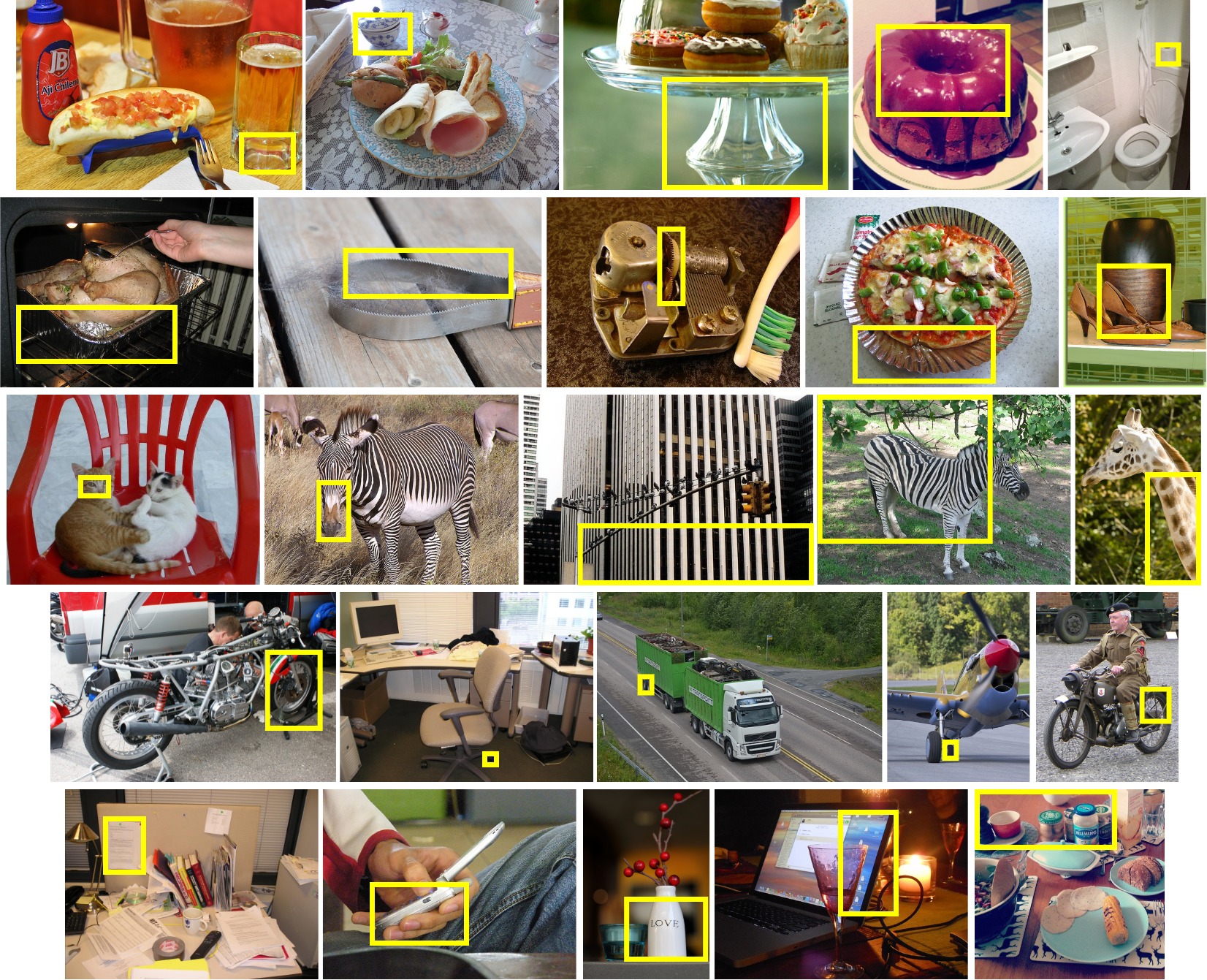}
\caption{Examples of semantics encoded in diverse prototypes. }
\label{visualization_proto}
\end{figure}

We first study what each individual prototype learns. In Figure \ref{visualization_proto}, we visualize instances (regions inside bounding boxes) most relevant to the prototypes (measured with their relevance scores on different prototypes, \textit{i.e.,} $\alpha_{p}^{k}$ in Equation \ref{sim_score_k}, $k$ denotes the indices of prototypes). The results show that our prototypes learn to represent a diverse pool of semantics. They not only capture low-level visual cues, such as shapes (\textit{e.g.,} round objects in the $1^{st}$ row), textures (\textit{e.g.,} objects with jagged texture in the $2^{nd}$ row), and patterns (\textit{e.g.,} objects with stripes in the $3^{rd}$ row), but also encode high-level semantics including object categories (\textit{e.g.,} wheels in the $4^{th}$ row) and commonalities in semantics (\textit{e.g.,} all objects in the $5^{th}$ row are displaying text). 

\begin{table}
\caption{Different groups of objects that are clustered based on their relevance to prototypes. Please refer to our supplementary materials for the complete results with 30 groups.}
\centering
\resizebox{1\linewidth}{!}{
{\small
\begin{tabular}{|c|c|}
\hline
Group & Objects \\
\hline
1& \makecell{cup, saucer, glass, beer, mug, \\  juice, beverage, liquid, smoothie, coffee} \\
\hline
2 & \makecell{fork, spoon, knife, silverware, utensil, \\ ladle, chopstick, tongs, spatula, butter knife} \\
\hline
3 & \makecell{spectator, umpire, batter, catcher, crowd, \\ net, player, baseball, stadium, bleachers} \\
\hline
4 & \makecell{bed, sofa, pillow, bedspread, headboard, \\ comforter, couch, sheet, ottoman, mattress} \\
\hline
5 & \makecell{sticker, newspaper, paper, sign, book, \\ tape, drawing, CD, letter, label}\\
\hline
6 & \makecell{water, sand, sea, rock, ocean, \\ boulders, lake, beach, shore, river} \\
\hline
\end{tabular}
}
}
\label{clustering_result}
\end{table} 

With our prototypes encoding abundant semantics, we further analyze their effectiveness in correlating relevant objects based on their characteristics. Specifically, we calculate the average relevance scores $\alpha_{p}$ (see Equation \ref{sim_score}) between objects in the GQA dataset and all prototypes, and then apply k-means algorithm \cite{kmeans} (k=30) to cluster objects using the scores. As shown in Table \ref{clustering_result}, by representing objects based on the prototypes, we can correlate objects that belong to similar categories (\textit{e.g.,} drinks and utensils in the $1^{st}$ and $2^{nd}$ rows), commonly appear in the same scenarios (\textit{e.g.,} baseball games and bedrooms in the $3^{rd}$ and $4^{th}$ rows) or share similar characteristics (\textit{e.g.,} objects related to text and landscape in the $5^{th}$ and $6^{th}$ rows). The results demonstrate the effectiveness of our prototypes in parsing objects based on their commonalities in the semantic space.

\subsection{How Do Models Reason to Answer Questions?} \label{reasoning_analysis}
Besides improving the generalizability, our method also enables interpretation of the decision-making process by visualizing the regions of interest (ROIs) at each reasoning step and prototypes matched with the observations. In this section, we provide qualitative examples to study the underlying rationales behind the derivation of answers.
\begin{figure}[t]
\centering
\includegraphics[width=1\linewidth]{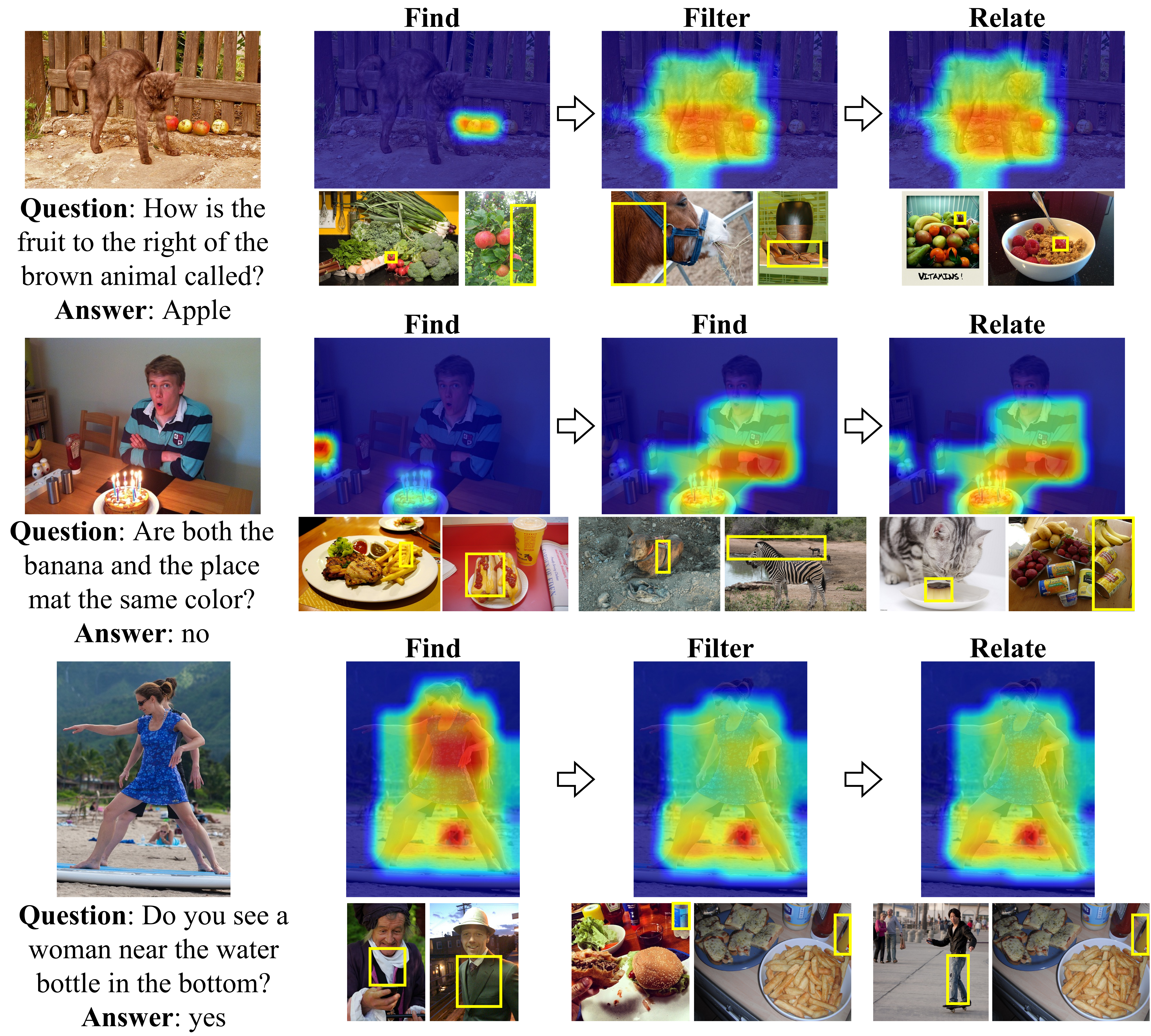}
\caption{Illustrations of the reasoning process. From left to right are the input images together with questions and predicted answers, and sequences of reasoning steps.}
\label{visualization_inference}
\end{figure}

Figure \ref{visualization_inference} shows the reasoning process of our method. For each question, we visualize the reasoning steps represented by neural modules (top), attention maps highlighting the ROIs (middle, $\alpha^{t}$ in Equation \ref{att_computation}), and images associated to prototypes matched with the observations (bottom). It shows that our method can correlate objects based on various characteristics and locate those important ones. In the $1^{st}$ example, while the apples are not explicitly mentioned in the question, our model correlates them with prototypes for different fruits (``fruit'' is a keyword in the question) and pays focused attention in the \textit{Find} step. Besides capturing semantic relationships about object categories, prototypes also help identify the cat based on its attribute (\textit{i.e.,} brown color) within the \textit{Filter} step, and enable the model to reason on the relative position between objects (\textit{i.e.,} \textit{Relate} step). In the $2^{nd}$ example, our model not only highlights the correct objects in the first two \textit{Find} steps (\textit{i.e.,} the banana and the mat), but also identifies the key characteristics that contribute to reasoning (\textit{i.e.,} matching observations with prototypes related to colors instead of object categories or other attributes). In the $3^{rd}$ example, woman in the question is an unseen object during training. With object factorization, our model successfully associates it to known objects (\textit{i.e.}, men) based on the similarity in clothing, and progressively shifts the attention from both women (\textit{i.e.,} \textit{Find} step) to the one lying next to the bottle (\textit{i.e.,} \textit{Filter} and \textit{Relate} steps).

\section{Conclusion}

This study is an effort toward generalizable and interpretable AI systems for real-world applications. It draws inspiration from the ways humans reason with the visual world, and investigates the effectiveness of integrating the compositionality of objects and the reasoning process. Our work distinguishes itself with a principled method for automatically factorizing objects into fine-grained semantics, bridging novel and known objects, and a new neural module network with a compositional decision-making process. The compositionality in both dimensions addresses object diversity and spurious data biases, enhancing model generalizability toward a broad range of scenarios. It also enables interpretation of the rationales behind the model's decisions. Experimental results demonstrate the advantages of our method under diverse settings, and provide insights on how our model reasons with the visual-textual inputs. We hope that this study can be useful for future developments of trustworthy visual reasoning models with more human-like intelligence and generalizability.

\section*{Acknowledgements}
This work is supported by NSF Grants 2143197 and 2227450.

{\small
\bibliographystyle{ieee_fullname}
\bibliography{egbib}
}

\clearpage

\section{Supplementary Materials}
Our supplementary materials consist of additional experimental results that demonstrate the effectiveness of the proposed method:
\begin{enumerate}
    \item We perform experiments to validate the generalizability of our method in two additional scenarios, \textit{i.e.,} questions about small or rare objects, respectively. (Section \ref{additional_exp}) 

    \item We present an ablation study on the model design of our prototypical neural module network. (Section \ref{ablation_model})

    \item We provide more detailed results on how our proposed prototypes characterize objects based on their semantic relationships (Section \ref{clustering_res}).

\end{enumerate}

\subsection{Additional Generalizability Experiments} \label{additional_exp}
One of the primary focuses of our study is to develop computational methods that can generalize toward various real-world scenarios. In the main paper, we show that, with object factorization and compositional reasoning, our approach is able to generalize to questions with novel concepts or variable data distributions. To further demonstrate its advantages, in this section we carry out experiments on two additional generalization settings: (1) questions with small objects, and (2) questions with rare objects. The former one considers ten novel objects with small sizes in the VQA dataset, \textit{e.g.,} bottle, bird, tie. While the latter reorganizes the GQA training and evaluation data by removing training questions with the top 30$\%$ least frequent objects and only considering evaluation questions with the removed objects. Results show that our method is advantageous in both scenarios, \textit{i.e.,} POEM: 52.11 and 55.78 vs XNM: 50.77 and 54.62. 

\subsection{Ablation Study on Model Design} \label{ablation_model}
The proposed prototypical neural module (POEM) network incorporates discriminative prototypes within different neural modules (\textit{i.e.,} \textit{Find}, \textit{Filter}, \textit{Describe}), and considers prototypes attended over time with a semantic memory module. In this section, we investigate the effectiveness of different components in the model by experimenting with two of its variants: (1) A variant that only utilizes prototypes in an independent \textit{Prototype} neural module (\textit{i.e.,} POEM-Ind), where \textit{Prototype} is an additional module alongside the existing ones in the XNM \cite{xnm} baseline, and (2) Our model without the semantic memory module (POEM w/o Mem). Experimental results in Table \ref{ablation_architecture} show that both variants lead to inferior performance, demonstrating the advantages of the design of our full POEM model.

\subsection{Additional Results on Object Characterization} \label{clustering_res}
In our main paper, we show that the proposed object factorization method is capable of learning semantically plausible prototypes. These prototypes encode abundant information about the characteristics of different objects, and can be used to cluster them into different groups with distinct semantic relationships (Table 4 of the main paper). This section provides the more detailed results of the clustering experiment, which contain the full list of objects divided into 30 groups. The results in Table \ref{clustering_result_supp} and Table \ref{clustering_result_supp_1} further demonstrate the usefulness of our prototypes for capturing the key characteristics of objects and correlating the relevant ones.
\begin{table}
\caption{Comparative results of our full model POEM and its variants. Best results are highlighted in bold.}
\centering
\resizebox{1\linewidth}{!}{
{\small
\begin{tabular}{cccccc}
\toprule
\multirow{2}{*}{} & \multicolumn{2}{c}{VQA \cite{vqa1}} & \multicolumn{2}{c}{GQA \cite{gqa}} & Novel-VQA \cite{skill_concept}\\
\cmidrule(lr){2-3} \cmidrule(lr){4-5} \cmidrule(lr){6-6}
 & Known & Novel & Known & Novel & Novel-Q\\ 
\midrule
POEM-Ind & 63.51 & 54.25 & 55.01 & 52.99 & 56.66 \\
\midrule
POEM w/o Mem & 62.49 & 53.89 & 58.39 & 58.60 & 59.41 \\
\midrule
POEM & \textbf{63.80} & \textbf{54.82} & \textbf{60.60} & \textbf{59.71} & \textbf{60.73} \\
\bottomrule
\end{tabular}
}
}
\label{ablation_architecture}
\end{table} 

\begin{table*}
\caption{Different groups of objects that are clustered based on their similarity with prototypes.}
\centering
\resizebox{0.85\textwidth}{!}{
{\small
\begin{tabular}{|c|c|}
\hline
Group & Objects \\
\hline
1& \makecell{cup, saucer, glass, beer, mug, juice, beverage, liquid, smoothie, coffee, \\ syrup, coffee cup, wine, milk, wineglass, soda, yogurt, tea, milkshake, ...} \\
\hline
2 & \makecell{fork, knives, spoon, knife, silverware, utensil, ladle, \\ chopstick, tongs, spatula, butter knife, scissors, whisk, sword, earbuds} \\
\hline
3 & \makecell{spectator, umpire, batter, catcher, crowd, net, player,\\ baseball, match, stadium, bleachers, dugout, team, soccer ball} \\
\hline
4 & \makecell{bed, pillowcase, sofa, pillow, bedspread, bedroom, headboard, \\ comforter, couch, sheet,  ottoman, armchair, mattress, bedding} \\
\hline
5 & \makecell{water, sand, swimming pool, sea, rock, ocean, boulders, lake, beach, shore, \\ river, cliff, ice, dock, stone, puddle, pond, mud, seaweed, canoe, harbor}  \\
\hline
6 & \makecell{vegetable, bowl, lettuce, salad, cabbage, pepper, broccoli, \\ carrot, soup, spinach, cereal, rice, dinner, asparagus, beans, pasta, basil, celery, garnish, ...} \\
\hline
7 & \makecell{pole, building, church, clock, roof, tower, barn, clock tower, house, \\ statue, bridge, cone, post, antenna, cross, flag, streetlight, tunnel, city, bricks, balcony, ...} \\
\hline
8 & \makecell{outlet, faucet, toilet, bathroom, towel, sink, shower, seat, lid, toilet paper, \\ dispenser, tissue, tiles, bathtub, drain, dryer, urinal, soap dish, paper towel, ...} \\
\hline
9 & \makecell{pants, shoe, sneakers, jeans, sandal, glove, snowboard, sock, snowboarder, boot, ski, \\ skateboarder, skateboard, snow pants, leggings, skater, vacuum, heel, shoelace, knee pad} \\
\hline
10 & \makecell{collar, hat, cap, scarf, backpack, t-shirt, shorts, glasses, pajamas, sweater, \\ jersey, uniform, sweatshirt, bracelet, dress, skirt, apron, robe, bandana, undershirt, cloths, ...} \\
\hline
11 & \makecell{bottle, can, vase, container, jar, candle, candle holder, canister, water bottle, \\ alcohol, saltshaker, spray can, dish soap, hand soap, bottle cap, shampoo, wineglasses, ...} \\
\hline
12 & \makecell{cake, crumbs, ice cream, muffin, coin, cupcake, candies, dessert, cream, donuts, sprinkles, \\ frosting, sponge, pastries, chocolate, sugar, balloon, icing, cookie, sour cream, sushi, ...} \\
\hline
13 & \makecell{table, chandelier, fireplace, window, ceiling, rug, curtain, chair, coffee table, floor, \\ door, fan, carpet, frame, side table, cabinet, mirror, entrance, shelf, dresser, mat, heater, ...} \\
\hline
14 & \makecell{cauliflower, onion, beef, meat, mushroom, egg, potatoes, chicken, tofu, pineapple, nut, \\ potato, zucchini, shrimp, sausage, pickles, peanut, coconut, corn, mozzarella, ...} \\
\hline
15 & \makecell{snow, ground, grass, field, fence, hill, meadow, zoo, dirt, forest, mountain, \\ yard, family, hillside, park, hay, plain, mound, farm, pasture, lawn, desert, fog} \\
\hline
16 & \makecell{cheese, bacon, napkin, dish, pizza, fish, food, butter, toast, tablecloth, sauce, sandwich, \\  topping, bread, ham, tray, gravy, biscuit, platter, breakfast, hamburger, hot dog, bun, ...} \\
\hline
17 & \makecell{tail, kite, bird, beak, seagull, sail, swan, duck, goose, feathers, parrot, \\ parachute, dolphin, pigeon, flamingo, eagle, shark, peacock, geese} \\
\hline
18 & \makecell{man, boy, girl, suit, lady, person, clothes, coat, men, child, snowsuit, \\ gentleman, people, passenger, guy, surfer, tourist, pedestrian, mother, athlete, ...} \\
\hline
19 & \makecell{tomato, strawberry, pepperoni, olive, berry, grape, blueberry, \\ raspberry, cherry, raisin, blackberry, beet, eggplant, meatballs} \\
\hline
20 & \makecell{airplane, propeller, airport, cockpit, aircraft, jet, \\ helicopter, terminal, shuttle, whale} \\
\hline
21 & \makecell{cat, figurine, doll, dog, panda, panda bear, bear, toy, nose, teddy bear, \\ eyes, monkey, lips, kitten, baby, rubber duck, polar bear, fur, pig, ...} \\
\hline 
22 & \makecell{sticker, newspaper, picture, paper, sign, number, book, tape, drawing, cd, paint, \\ package, letter, arrow, label, magazine, logo, display, painting, poster, word, ...} \\
\hline
23 & \makecell{orange, grapefruit, fruit, banana, watermelon, kiwi, apple, lemon, lime, cucumber, \\ produce, papaya, squash, pear, pumpkin, melon, avocado, mango, pomegranate, ...} \\
\hline
24 & \makecell{motorcycle, bike, tire, bicycle, train, car, minivan, cart, vehicle, van, truck, taxi, \\ train car, locomotive, SUV, firetruck, tractor, scooter, gas station, school bus, ...} \\
\hline
\end{tabular}
}
}
\label{clustering_result_supp}
\end{table*} 

\begin{table*}
\caption{Different groups of objects that are clustered based on their similarity with prototypes (cont'd).}
\centering
\resizebox{0.85\textwidth}{!}{
{\small
\begin{tabular}{|c|c|}
\hline
Group & Objects \\
\hline
25 & \makecell{pipe, pen, wire, cord, stick, toothbrush, chain, rope, brush, cable, straw, microphone, hose, \\ guitar, pencil, bat, tool, comb, instrument, gun, toothbrushes, steering wheel, ...} \\
\hline
26 & \makecell{elephant, giraffe, cow, calf, animal, zebra, sheep, donkey, lion, lamb, \\ goat, bull, herd, bison, wool, deer, antelope, ostrich, rhino, moose} \\
\hline
27 & \makecell{bush, flower, trunk, plant, leaf, branch, horn, tree, bushes, branches, leaves, pine tree, \\ feeder, palm tree, garden, rose, mane, daisy, bouquet, garland, ...} \\
\hline
28 & \makecell{box, drawer, microwave, counter, dishwasher, oven, stove, kitchen, bucket, sack, bag, \\ trashcan, toaster, burner, appliance, suitcase, refrigerator, pot, teapot, basket, ...} \\
\hline
29 & \makecell{binder, desk, screen, keyboard, laptop, television, phone, notebook, controller, computer, \\ monitor, headphones, mouse pad, printer, speaker, camera, game, ...} \\
\hline
30 & \makecell{street, path, platform, train station, walkway, pavement, road, gravel, sidewalk, runway, \\ railroad, parking lot, crosswalk, skate park, station, highway, intersection, barrier, ...} \\
\hline
\end{tabular}
}
}
\label{clustering_result_supp_1}
\end{table*}

\end{document}